%% file: main.tex
\documentclass{article} 
\usepackage{iclr2027_conference,times}

\input{math_commands.tex}

\usepackage{hyperref}
\usepackage{url}
\usepackage{amsmath,amssymb}
\usepackage{algorithm}
\usepackage{algpseudocode}
\usepackage{booktabs}
\usepackage{tabularx}
\usepackage{array}
\usepackage{graphicx}
\usepackage{wrapfig}
\usepackage{adjustbox}
\usepackage{caption}

\newcommand{\methodname}{\textbf{\texttt{CoEvolve}}}

\title{CoEvolve: Self-Evolving Code Policies from Skills to Strategies for Humanoid Robots}

\author{
\textbf{Zicheng Liu}$^{1}$,
\textbf{Haochen Wu}$^{2}$,
\textbf{Jiawei He}$^{3}$,
\textbf{Suyun Zhao}$^{2}$,
\textbf{Huan Yang}$^{4}$,\\
\textbf{Cheng Chi}$^{2,\dagger}$,
\textbf{Hao Tang}$^{1,\dagger}$ \\[5pt]
$^{1}$Peking University \qquad
$^{2}$Renmin University of China \\
$^{3}$Beijing XYZ Embodied AI Co., Ltd. \\
$^{4}$Beijing Academy of Artificial Intelligence (BAAI)
}

\hypersetup{
    pdftitle={
        CoEvolve: Self-Evolving Code Policies
        from Skills to Strategies for Humanoid Robots
    },
    pdfauthor={
        Zicheng Liu, Haochen Wu, Jiawei He,
        Suyun Zhao, Huan Yang, Cheng Chi, Hao Tang
    }
}

\iclrfinalcopy 
\begin{document}

\maketitle
\lhead{Preprint}

\begingroup
\renewcommand{\thefootnote}{\ensuremath{\dagger}}
\footnotetext{%
\raggedright
Corresponding authors:
\href{mailto:chengchi@ruc.edu.cn}
{\texttt{chengchi@ruc.edu.cn}},
\href{mailto:bjdxtanghao@gmail.com}
{\texttt{bjdxtanghao@gmail.com}}.
\par
}
\endgroup

\input{sections/0_abstract}
\input{sections/1_introduction}
\input{sections/related_work}
\input{sections/3_methodology}

\input{sections/4_experiments}
\input{sections/5_conclusions}

\bibliography{main}
\bibliographystyle{iclr2027_conference}


\newpage
\appendix
\onecolumn 
\input{sections/appendix}

\end{document}

%% file: math_commands.tex
\usepackage{amsmath,amsfonts,bm}

\def\eqref#1{equation~\ref{#1}}

\def\1{\bm{1}}

\DeclareMathAlphabet{\mathsfit}{\encodingdefault}{\sfdefault}{m}{sl}
\SetMathAlphabet{\mathsfit}{bold}{\encodingdefault}{\sfdefault}{bx}{n}



%% file: sections/0_abstract.tex
\begin{abstract}
Recent advances in Vision-Language-Action (VLA) models and World Action
Models (WAMs) have expanded the manipulation and whole body capabilities of
individual robots. Yet a single robot remains limited by its embodiment, and
many real world tasks require capabilities that can only emerge through
cooperation. Existing multi robot frameworks largely emphasize task
decomposition, allocation, and replanning, while providing limited support
for adapting executable coordination when physical execution deviates from
the intended behavior. We present \methodname{}, a framework that represents
team coordination as executable code over fixed robot motor skills.
\methodname{} introduces Verifier Guided Coordination Evolution, where a
multimodal foundation model instantiates coordination strategies through
motion primitives, semantic objectives, temporary robot roles, and bounded
execution parameters. Physical verification evaluates each executed
transition, rejected behaviors trigger structured coordination revision from
an accepted checkpoint, and verified experience is retained to guide
subsequent decisions. We evaluate \methodname{} with two simulated Unitree G1
humanoids transporting a freely supported rigid payload through constrained
corridors. Across 15 base task runs with three multimodal backbone families,
\methodname{} completes 8 tasks, with every backbone achieving end to end
success. Additional evaluations yield verified placement on two additional
corridor layouts and after both same-backbone and cross-backbone experience
initialization, while exposing persistent limitations in motion precision and
contact stability. These results support executable coordination as a
practical interface for adapting long horizon multi robot collaboration under
shared physical constraints. Our code will be available at
\url{https://github.com/Innovationlzc/CoEvolve}.
\end{abstract}

%% file: sections/1_introduction.tex
\section{Introduction}
\label{sec:introduction}

Recent advances in robot foundation models have equipped individual robots
with increasingly general visuomotor capabilities, enabling them to follow
language instructions and perform diverse manipulation
tasks~\citep{brohan2022rt,brohan2023rt,kim2024openvla,black2024pi_0}.
Yet some physical tasks require capabilities that cannot be provided by one
robot alone. Transporting an extended object through a constrained environment, for
example, requires coordinated support, relative motion, corner negotiation,
and placement~\citep{de2023centralized,pandit2025multibiped}.
Collaboration is therefore not simply the aggregation of individual skills.
Robots that can independently walk, turn, and move laterally must still
determine how to combine these capabilities while carrying a shared object.
We view this formation of team behavior as a form of \emph{co-becoming}:
effective cooperation is realized through the changing relationship between
individual capabilities and the shared physical state.

\begin{figure}[t]
    \centering
    \includegraphics[width=\linewidth]{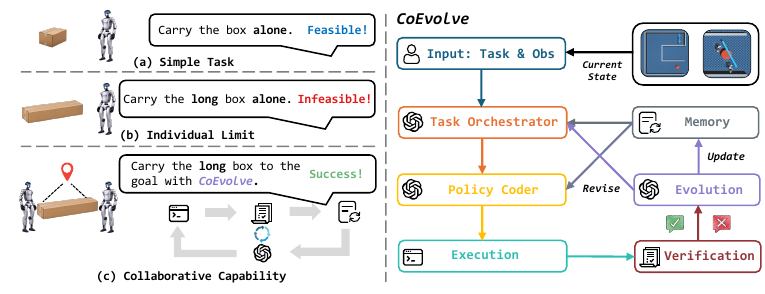}
    \vspace{-5mm}
    \caption{
    \textbf{Motivation and overview of \methodname{}.}
    The schematic contrasts individual manipulation with coordinated
    transport of a shared payload. \methodname{} combines executable team
    policies, physical verification, and reusable experience to revise
    coordination during execution.
    }
    \label{fig:teaser}
    \vspace{-5mm}
\end{figure}

This motivates our central question: \emph{given fixed individual motor
skills, how can an agent instantiate and revise executable team coordination
as a shared physical task unfolds?}
Multi-robot planning methods address task decomposition, allocation, and
collaboration through structured and language-model planning
~\citep{cai2025mrbtp,mandi2024roco,kannan2024smart,
liu2025coherent,yuan2025remac},
while learned coordination methods optimize joint or decentralized policies
through training~\citep{zhang2026halo,doshi2026chorus}. In parallel, agents improve
textual skills, memory systems, and executable implementations through
feedback and experience
sharing~\citep{yang2026skillopt,zhang2025memevolve,weng2026gea}.
Code as Policies and subsequent robot programming systems also generate and
revise executable behaviors through
interaction~\citep{liang2023code,fu2026capx,lu2026aspire,xiao2026enpire}.
We study a complementary challenge: revising coordination when robots are
physically coupled through a shared payload. In this setting, an individual
action can alter both a teammate's support conditions and the feasibility of
subsequent joint motion. As a result, a valid task plan does not guarantee
that the current coordination pattern remains executable. The key challenge
is therefore to instantiate joint behavior from fixed motor skills and revise
the relevant coordination relation when physical execution deviates, for
example when unequal displacement disrupts formation or a constrained turn
requires temporary support and mover roles.

We introduce \methodname{}, a framework for adapting executable team
coordination during shared physical tasks. Figure~\ref{fig:teaser} summarizes
the motivation and overall interaction loop. \methodname{} separates fixed
individual motor capabilities from the coordination strategy that composes
them for the current task phase. Rather than selecting from a library of
complete strategies, the agent instantiates coordination through motion
primitives, semantic objectives, temporary roles, coordination modes, and
bounded execution parameters. The evaluated setting provides simulator
measurements and route references, while the model determines how these
capabilities should be combined within the resulting physical constraints.

Our Verifier Guided Coordination Evolution mechanism uses physical execution
evidence to determine which aspects of coordination should change. A rejected
transition triggers a revision to the maneuver, the relationship between
robot roles, or the execution parameters. Revised policies are evaluated
after restoration of the latest accepted simulator checkpoint. Verified
outcomes also contribute complementary forms of experience: abstract
coordination records guide subsequent decisions, while accepted executable
examples inform code generation. Foundation model and motor policy weights
remain fixed. The adaptation concerns the task plan, instantiated
coordination strategies, and stored experience rather than the acquisition
of new motor skills.

We evaluate \methodname{} with two simulated Unitree G1 humanoids carrying a
freely supported rigid payload through constrained corridors. The base
evaluation contains eight complete tasks across fifteen runs and three
foundation model families. We further compare against four adapted external
baselines and two component variants, evaluate additional corridor layouts,
and study reuse of verified cooperative experience across foundation model
backbones. These experiments distinguish demonstrated execution capability
from causal component effects: some comparison, scene, and transfer records
differ in runtime configuration, interaction budget, or continuation
conditions. Our contributions are:

\begin{itemize}
    \item We introduce an executable representation of team coordination
    that separates fixed individual motor capabilities from the cooperative
    strategy instantiated for the current physical situation.

    \item We propose Verifier Guided Coordination Evolution, which combines
    physical verification, structured strategy revision, and reusable
    cooperative experience to adapt coordination during execution.

    \item We provide an extensive evaluation across three foundation model
    families, adapted multi robot and code policy baselines, additional corridor
    layouts, and cross-model experience reuse, while analyzing the execution and
    contact limitations that remain.
\end{itemize}

\begin{figure}[t]
    \centering
    \includegraphics[width=\linewidth]{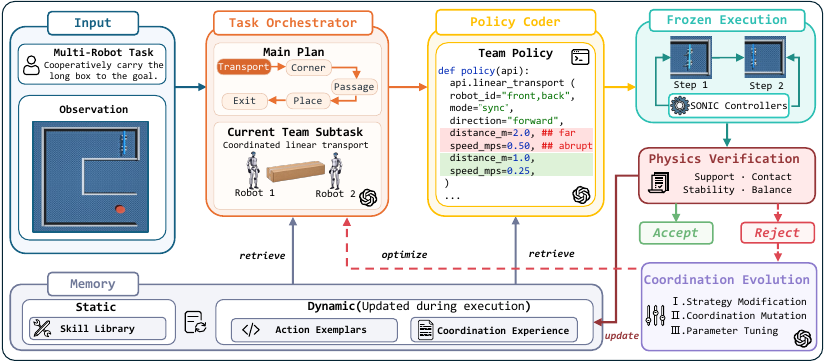}
    \caption{
    \textbf{Framework of \methodname{}.}
    The Task Orchestrator commits a team subtask, which the Policy Coder
    grounds into executable robot policies. Physical verification accepts or
    rejects the resulting transition; rejected executions trigger structured
    coordination revision, while verified outcomes update cooperative
    experience. Motor-policy weights remain frozen.
    }
    \label{fig:framework}
\end{figure}

%% file: sections/related_work.tex
\section{Related Work}
\label{sec:related_work}

\paragraph{Self-Evolving Agents.}
Self-evolving agents adapt multiple components of an agent system.
Building on feedback accumulation and skill reuse in Reflexion and
Voyager~\citep{shinn2023reflexion,wang2024voyager}, SkillOpt optimizes textual
skills for frozen agents, while MemEvolve evolves the memory architecture
itself~\citep{yang2026skillopt,zhang2025memevolve}. Continual Harness adapts
prompts, sub-agents, skills, and memory within an ongoing episode, and
Hyperagents additionally makes the improvement procedure
editable~\citep{karten2026continualharness,zhang2026hyperagents}. Related systems automate the design or modification of the agent itself,
including Automated Design of Agentic Systems and the Darwin G\"odel
Machine~\citep{hu2024adas,zhang2025dgm}. Group-Evolving Agents (GEA) enables experience sharing among evolving coding
agents, whereas CORAL supports autonomous multi-agent search through shared
persistent memory~\citep{weng2026gea,qu2026coral}. These approaches concern
agent implementations, knowledge, or search processes. \methodname{} instead
studies online revision of executable inter-robot coordination during a
shared physical task, with motor skills and runtime contracts fixed.

\paragraph{Robot Policy Refinement from Execution Feedback.}
REMAC revises multi-robot task plans through reflection, while HyCodePolicy
and Act-Observe-Rewrite revise executable robot programs from execution
feedback~\citep{yuan2025remac,liu2025hycodepolicy,kumar2026aor}.
ASPIRE combines multimodal failure diagnosis, program repair, and reusable
skill discovery, including bimanual manipulation; ENPIRE automates policy
improvement through real-robot experimentation and
reset~\citep{lu2026aspire,xiao2026enpire}. These works already connect
program revision with physical experience. \methodname{} focuses on
physically coupled humanoid transport, where failures can require changes
to the selected maneuver or temporary support roles. It combines structured
coordination revision with physical transition checks, checkpoint recovery,
and sequential retrieval of strategy records and accepted code exemplars.

%% file: sections/3_methodology.tex
\section{Method}
\label{sec:method}

\begin{algorithm}[t]
\caption{\methodname{}: Verifier Guided Coordination Evolution}
\label{alg:coevolve}
\small
\begin{algorithmic}[1]

\Require Task $g$, capability interface $\mathcal{K}$,
initial experience $\mathcal{M}_0$, budget $B$

\State $P\gets\Call{InitializePlan}{g}$;
$\mathcal{M}\gets\mathcal{M}_0$;
$\bar{s}\gets s_0$;
$\xi\gets\varnothing$
\State $C\gets\Call{SaveCheckpoint}{\bar{s}}$

\For{$t=0,\ldots,B-1$}

    \State $o_t\gets\Call{Observe}{\bar{s}}$;
    $q_t\gets\Call{ActivePhase}{P}$

    \State $r_t^D\gets
    \Call{RetrieveStrategy}{\mathcal{M}^D,q_t,\xi}$;
    $z_t\gets\Call{Orchestrate}{g,o_t,P,r_t^D}$

    \State $r_t^A\gets
    \Call{RetrieveExemplar}{\mathcal{M}^A,z_t}$;
    $(\chi_t,\Pi_t)\gets
    \Call{GeneratePolicy}{z_t,o_t,r_t^A,\mathcal{K}}$

    \State $s_t^+\gets\Call{Execute}{\bar{s},\Pi_t}$;
    $E_t\gets\Call{Verify}{\bar{s},\Pi_t,s_t^+}$

    \State $\mathcal{M}\gets
    \Call{UpdateExperience}{\mathcal{M},\chi_t,\Pi_t,E_t}$

    \If{$E_t^{\mathrm{acc}}$}
        \State $\bar{s}\gets s_t^+$;
        $C\gets\Call{SaveCheckpoint}{\bar{s}}$;
        $P\gets\Call{AdvanceIfComplete}{P,E_t}$;
        $\xi\gets\varnothing$

        \If{$\mathcal{S}(\bar{s},g)$}
            \State \Return success
        \EndIf

    \Else
        \State \Call{Restore}{$C$};
        $\xi\gets\Call{FailureContext}{E_t,\chi_t}$
        \State $\ell_t\gets
        \Call{SelectRevisionScope}{E_t,\chi_t}$;
        $\chi_t'\gets
        \Call{ReviseCoordination}{\chi_t,\ell_t,E_t}$
        \State \Call{CommitRevision}{$\chi_t'$}
        \Comment{constrains next orchestration and policy generation}
    \EndIf

\EndFor

\State \Return failure

\end{algorithmic}
\end{algorithm}

Figure~\ref{fig:framework} illustrates the overall architecture of
\methodname{}, from semantic coordination decisions to executable policies,
physical verification, and experience-driven revision.

\subsection{Overview and Coordination Representation}
\label{sec:method_overview}
\label{sec:coordination_strategy}

\paragraph{Task formulation.}
\methodname{} separates fixed individual motor skills from mutable team
coordination. For $N$ robots manipulating a shared payload, a persistent
semantic plan $P_t=[(q_k,\gamma_k,\sigma_{k,t})]_{k=1}^{K}$ records task
phases, completion criteria, and status. The host provides route-segment
geometry and progress constraints, while the model determines how the robots
coordinate within those constraints. State, observation, and route interfaces
are detailed in Appendix~\ref{app:state_and_roles}.

\paragraph{Executable coordination.}
We use executable code as the coordination interface, following the
Code as Policies paradigm~\citep{liang2023code}, but generate paired robot
policies under a shared coordination decision. Above the fixed motor
interface, \methodname{} exposes coordinated linear transport, pivot, and
alignment. The Task Orchestrator commits
$z_t=(\kappa_t,d_t,m_t,\rho_t^{1:N})$, where $\kappa_t$ is the coordination
primitive, $d_t$ the semantic objective,
$m_t\in\{\texttt{sync},\texttt{asym}\}$ the coordination mode, and
$\rho_t^{1:N}$ the temporary robot roles. These variables denote the
coordination fields of the committed subtask; its structured representation
also carries formation intent, completion evidence, and execution constraints.

The Policy Coder supplies bounded execution parameters $\theta_t^{1:N}$,
yielding $\chi_t=(z_t,\theta_t^{1:N})$ and executable team policy
$\Pi_t=(\pi_t^1,\ldots,\pi_t^N)$. Each robot receives a separate policy
containing one allowed action call followed by an explicit stop. The host
validates the paired policies as one logical team transaction and dispatches
them to separate motor deployments. Logical grouping does not imply identical
physical response. The motor interface and execution constraints are given in
Appendix~\ref{app:skills_and_primitives}.

\subsection{Verifier Guided Coordination Evolution}
\label{sec:coordination_evolution}

Executing $\Pi_t$ from the latest accepted state $s_t$ produces candidate
state $s_t^+$. The verifier returns
$E_t=\operatorname{Verify}(s_t,\Pi_t,s_t^+)$ and determines acceptance from
measured task progress, payload support and stability, robot configuration,
and environmental contacts. A multimodal evaluator assists scene
interpretation and failure diagnosis, while measured physical quantities
determine transition acceptance. An accepted transition becomes the next
trusted checkpoint; the semantic plan advances only when the active phase
satisfies its completion criterion.

A rejected transition triggers typed revision rather than unrestricted policy
regeneration. Given failed strategy $\chi_t$ and verifier evidence $E_t$, the
revision scope
$\ell_t\in\{\textsc{Strategy},\textsc{Coordination},\textsc{Parameter}\}$
changes $(\kappa_t,d_t)$, $(m_t,\rho_t^{1:N})$, or
$\theta_t^{1:N}$, respectively. Fields outside the selected scope remain
fixed. The approved child strategy constrains the next policy-generation step
rather than being replaced by an unrelated coordination decision, and the
resulting policy is evaluated again from the latest accepted checkpoint.
Verification, revision validation, and recovery rules are detailed in
Appendix~\ref{app:verification_and_revision}.

\subsection{Verified Cooperative Experience}
\label{sec:cooperative_memory}

\methodname{} maintains two dynamic stores,
$\mathcal{M}_t=(\mathcal{M}_t^D,\mathcal{M}_t^A)$. Strategy memory
$\mathcal{M}_t^D$ stores abstract coordination knowledge, while action
exemplar memory $\mathcal{M}_t^A$ stores accepted executable implementations.
Strategy records capture applicability conditions, observed failure patterns,
and repair directions, whereas action exemplars retain executable code and
effective parameters.

Before orchestration, strategy records are retrieved using the active phase
$q_t$ and recent failure context $\xi_t$ as
$r_t^D=\operatorname{RetrieveStrategy}(\mathcal{M}_t^D;q_t,\xi_t)$.
After the Orchestrator commits $(\kappa_t,d_t,m_t)$, compatible action
exemplars are retrieved as
$r_t^A=\operatorname{RetrieveExemplar}
(\mathcal{M}_t^A;\kappa_t,d_t,m_t)$.
Thus, strategy retrieval informs \emph{what coordination to attempt}, while
exemplar retrieval informs \emph{how to implement the selected coordination}.

Experience is updated only after physical execution:
\[
\mathcal{M}_{t+1}
=
\operatorname{UpdateExperience}
(\mathcal{M}_t,\chi_t,\Pi_t,E_t).
\]
Accepted transitions may provide reusable strategy records and executable
exemplars, while rejected transitions may provide verified counterexamples.
Unexecuted repair proposals are not promoted as successful experience.
Record schemas, retrieval, and promotion rules are given in
Appendix~\ref{app:memory_retrieval}.

%% file: sections/4_experiments.tex
\section{Experiments}
\label{sec:experiments}

\begin{table}[t]
    \centering
    \small
    \setlength{\tabcolsep}{5pt}
    \renewcommand{\arraystretch}{1.10}
    \begin{tabular*}{\linewidth}{
        @{\extracolsep{\fill}}lcccc@{}
    }
        \toprule
        \textbf{Backbone}
        & \shortstack{\textbf{Full-task}\\\textbf{SR@80} $\uparrow$}
        & \shortstack{\textbf{VP}\\\textbf{(\%)} $\uparrow$}
        & \shortstack{\textbf{EA}\\\textbf{(\%)} $\uparrow$}
        & \shortstack{\textbf{PAUC@80}\\\textbf{(\%)} $\uparrow$} \\
        \midrule

        GPT-5.4-mini
        & \textbf{3/5 (60.0\%)}
        & \textbf{86.7}
        & 85.4
        & 61.9 \\

        Gemini-3.5-Flash
        & \textbf{3/5 (60.0\%)}
        & 76.7
        & 66.7
        & 60.2 \\

        Claude-Sonnet-4.6
        & 2/5 (40.0\%)
        & 76.7
        & \textbf{86.5}
        & \textbf{64.5} \\

        \bottomrule
    \end{tabular*}
    \caption{
    Base-task results across foundation model backbones
    ($n=5$ per backbone). Success requires placement, release, and robot
    retreat. VP, EA, and PAUC are per-run means.
    }
    \label{tab:backbone_main}
\end{table}

\subsection{Experimental Setting}
\label{sec:exp_setup}

\paragraph{Task and observations.}
We evaluate \methodname{} on cooperative payload transport in
MuJoCo~\citep{todorov2012mujoco}. Two Unitree G1 humanoids jointly carry a
freely supported $0.5\,\mathrm{kg}$ rigid payload through constrained
corridors. The task requires coordinated transport, corner negotiation,
target placement, and release. The base task additionally requires both
robots to retreat while remaining upright.

The model receives simulator measurements, global and payload-centered
observations, the active task phase, and retrieved experience when enabled.
The host provides route-segment geometry and progress constraints. The model
selects the coordination primitive, temporary roles, coordination mode, and
bounded execution parameters within these constraints.

\paragraph{Execution and models.}
Each robot uses an independent SONIC deployment with frozen policy
weights~\citep{luo2025sonic}, together with task-specific skill wrappers and
support posture control. The host validates paired robot policies as one team
transaction and dispatches their planner commands. The controllers retain
independent observation histories and execution threads, so joint dispatch
does not imply identical physical responses.

We evaluate GPT-5.4-mini, Gemini-3.5-Flash, and Claude-Sonnet-4.6. One
backbone performs all semantic roles within a run. Except for experience
reuse experiments, dynamic memory starts empty and updates remain local to
that run.

\subsection{Evaluation Protocol and Metrics}
\label{sec:exp_protocol}
\label{sec:metrics}

A \emph{physical transaction} is one joint robot command dispatched for
physical execution. Accepted and rejected executions consume the physical
budget, while outputs rejected before dispatch consume model calls but no
physical transactions. The base task uses an 80-transaction budget and
requires placement, release, and robot retreat. Layout and experience-reuse
evaluations use verified placement and release as their endpoint.

We report success rate (SR), verified progress (VP), execution acceptance
(EA), and Progress AUC (PAUC). SR measures completion of the relevant
endpoint. VP is
$\mathrm{VP}=100n^{-1}\sum_r q_r/K_r$, where $q_r$ of the $K_r$
scene-defined milestones are certified at the final trusted checkpoint.
EA is
$\mathrm{EA}=100n^{-1}\sum_r A_r/T_r$, the mean per-run fraction of
dispatched transactions accepted by the verifier. Unresolved charged outcomes
remain in the denominator and produce the confirmed lower bounds marked by
$\geq$. EA does not include failures that occur before physical dispatch.

PAUC measures when verified milestones are first acquired. Let
$\tau_{rk}$ denote the first cumulative physical transaction at which
milestone $k$ is certified, with $\tau_{rk}=\infty$ if it is never certified.
For reporting horizon $B$,
\[
\mathrm{PAUC@}B
=
\frac{100}{n}
\sum_{r=1}^{n}
\frac{1}{K_rB}
\sum_{k=1}^{K_r}
\left[B-\min(\tau_{rk},B)\right].
\]
Rejected transactions remain charged, rollback does not refund interaction,
and each milestone contributes once. After termination, certified progress is
held constant until $B$. PAUC therefore measures progress attainment and its
timing, not terminal stability or inference efficiency. VP measures progress retained at the final trusted checkpoint, whereas PAUC
measures when verified milestones are first attained.

We use $B=80$ for base comparisons, a retrospective horizon of $B=160$ for
additional layouts, and the first 80 cumulative transactions for transfer
PAUC. Other transfer metrics describe the complete selected execution.
Physical transactions, foundation-model calls, terminal categories, and
run-time accounting are reported in
Appendices~\ref{app:execution_diagnostics} and~\ref{app:runtime}.
The full comparison condition reuses the GPT backbone runs, and differences
in runtime and continuation settings make the aggregate comparisons
descriptive rather than matched causal estimates.

\begin{table*}[t]
    \centering
    \footnotesize
    \setlength{\tabcolsep}{5pt}
    \renewcommand{\arraystretch}{1.10}
    \begin{tabularx}{\textwidth}{
        @{}>{\raggedright\arraybackslash}Xcccc@{}
    }
        \toprule
        \textbf{Method}
        & \shortstack{\textbf{Full-task}\\\textbf{SR@80} $\uparrow$}
        & \shortstack{\textbf{VP}\\\textbf{(\%)} $\uparrow$}
        & \shortstack{\textbf{EA}\\\textbf{(\%)} $\uparrow$}
        & \shortstack{\textbf{PAUC@80}\\\textbf{(\%)} $\uparrow$} \\
        \midrule

        \multicolumn{5}{@{}l}{\emph{External baselines}} \\
        \addlinespace[2pt]

        REMAC~\citep{yuan2025remac}
        & 0/5 (0.0\%) & 26.7 & 57.1 & 23.3 \\

        COHERENT~\citep{liu2025coherent}
        & 0/5 (0.0\%) & 56.7 & $\geq 83.6$ & 49.1 \\

        CaP-Agent0~\citep{fu2026capx}
        & 0/5 (0.0\%) & 33.3 & 66.7 & 29.5 \\

        ASPIRE (Online)~\citep{lu2026aspire}
        & 0/5 (0.0\%) & 40.0 & 74.9 & 35.6 \\

        \addlinespace[3pt]
        \multicolumn{5}{@{}l}{\emph{Component variants}} \\
        \addlinespace[2pt]

        Flat Reflective Repair
        & 0/5 (0.0\%) & 43.3 & 73.2 & 36.3 \\

        \methodname{} w/o Dynamic Experience
        & 0/5 (0.0\%) & 60.0 & 63.8 & 46.4 \\

        \midrule

        \textbf{\methodname{}}
        & \textbf{3/5 (60.0\%)}
        & \textbf{86.7}
        & \textbf{85.4}
        & \textbf{61.9} \\

        \bottomrule
    \end{tabularx}
    \caption{
    Base-task comparison using GPT-5.4-mini ($n=5$ per condition).
    The full method reuses the GPT runs in
    Table~\ref{tab:backbone_main}. $\geq$ denotes a confirmed EA lower bound.
    }
    \label{tab:baselines_ablations}
\end{table*}

\subsection{Cross-Backbone Evaluation}
\label{sec:backbone_results}

We first test whether the same coordination interface supports different
foundation model backbones. Each backbone is evaluated on five base-task runs
under the same task definition and endpoint.

\paragraph{Local acceptance does not guarantee sustained task advancement.}
Table~\ref{tab:backbone_main} shows successful full-task execution under all
three backbones. Claude achieves the highest EA (86.5\%) and PAUC@80
(64.5\%) but fewer completions than GPT and Gemini, separating local
execution validity from terminal success. The same distinction appears
within one GPT run: after reaching the fourth milestone at transaction 47,
31 of the remaining 33 transactions are accepted, including 27 setup actions,
yet no further milestone is certified. Claude's failed runs instead terminate
during generation or repair, which EA does not measure because no physical
transaction is dispatched. Long-horizon performance therefore depends on
revisions that continue phase advancement, not only on verifier-accepted
execution.

\subsection{Method Comparisons and Component Analysis}
\label{sec:baseline_ablation}

We next test whether physical feedback alone is sufficient for long-horizon
repair, and whether structured revision and verified experience address
different failure modes. We compare \methodname{} with
REMAC~\citep{yuan2025remac}, COHERENT~\citep{liu2025coherent},
CaP-Agent0~\citep{fu2026capx}, and ASPIRE~\citep{lu2026aspire}, adapted to
the same cooperative transport task and robot interface. We also evaluate
Flat Reflective Repair and \methodname{} without Dynamic Experience.
Implementation details are given in
Appendix~\ref{app:baseline_configurations}.

\begin{figure*}[t]
    \centering
    \includegraphics[width=\textwidth]{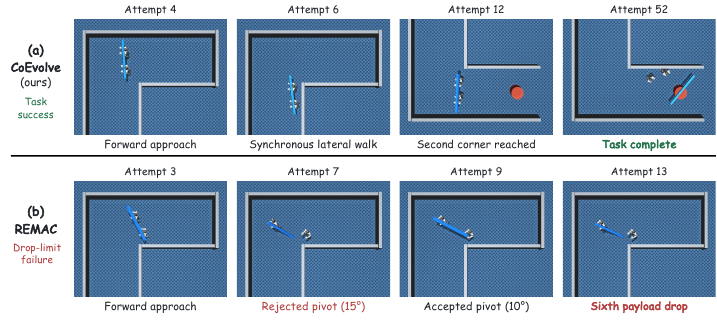}
    \vspace{-5mm}
    \caption{
    \textbf{Representative CoEvolve and REMAC executions on the base task.}
    CoEvolve reaches task completion, whereas REMAC terminates at the
    cumulative payload-drop limit after successive pivot attempts.
    Attempt labels denote internal execution indices rather than cumulative
    physical transactions.
    }
    \label{fig:baseline_vs_ours}
    \vspace{-2mm}
\end{figure*}

\paragraph{Structured revision and verified experience target different repair failures.}
Table~\ref{tab:baselines_ablations} shows full-task completion only for
\methodname{}, although all alternatives reach partial progress. A
CaP-Agent0 trace illustrates why feedback alone can be insufficient: after
receiving a requirement for different yaw corrections, the repair changes
both values from 2.5 to 2.0 and preserves the rejected relation. Typed
revision instead separates changes to the maneuver, role relation, and
execution parameters. The component variants expose a second distinction.
Removing Dynamic Experience still reaches 60.0\% VP but yields no completion,
with EA at 63.8\% and PAUC@80 at 46.4\%. Flat Reflective Repair terminates at
the payload-drop limit in four runs, whereas the no-memory variant ends in
four generation/repair stalls and one budget exhaustion. These failure
profiles are consistent with structured revision localizing what should
change and verified experience supplying evidence for what to try next,
without isolating either effect causally.

Figure~\ref{fig:baseline_vs_ours} provides a concrete execution-level view of
the contrast observed in Table~\ref{tab:baselines_ablations}. Both methods
reach a nontrivial turning stage, but their subsequent executions diverge.
\methodname{} continues through the corner and eventually completes the task,
whereas REMAC first produces a rejected pivot, then executes a smaller
accepted pivot, but later terminates at the payload-drop limit. The comparison
shows that local executability and verifier acceptance are not sufficient for
sustained task progress: an accepted correction may still leave the team in a
configuration that does not support the remaining transport sequence. This
trace is consistent with the role of structured revision in changing the
coordination relation after failure, rather than only producing another
locally executable motion.

\subsection{Evaluation Across Layouts}
\label{sec:scene_generalization}

We then test whether successful coordination on the base task remains
executable under different corridor geometries. Three additional layouts are
evaluated with GPT-5.4-mini using three runs per layout. The endpoint is
verified placement and release without the final robot retreat. Scene geometry
is shown in Appendix~\ref{app:generalization_scenes}.

\begin{table}[t]
    \centering
    \small
    \setlength{\tabcolsep}{6pt}
    \renewcommand{\arraystretch}{1.10}
    \begin{tabular*}{\linewidth}{
        @{\extracolsep{\fill}}lcccc@{}
    }
        \toprule
        \textbf{Layout}
        & \shortstack{\textbf{Verified}\\\textbf{placement} $\uparrow$}
        & \shortstack{\textbf{VP}\\\textbf{(\%)} $\uparrow$}
        & \shortstack{\textbf{EA}\\\textbf{(\%)} $\uparrow$}
        & \shortstack{\textbf{PAUC@160}\\\textbf{(\%)} $\uparrow$} \\
        \midrule

        Layout 2
        & 1/3 (33.3\%)
        & 41.7
        & 71.4
        & 30.3 \\

        Layout 3
        & 0/3 (0.0\%)
        & 58.3
        & 46.1
        & 53.6 \\

        Layout 4
        & 1/3 (33.3\%)
        & 50.0
        & 64.1
        & 37.1 \\

        \bottomrule
    \end{tabular*}
    \caption{
    Additional-layout results ($n=3$ per layout).
    PAUC@160 uses a common retrospective reporting horizon.
    }
    \label{tab:scene_generalization}
\end{table}

\begin{figure}[t]
    \centering
    \includegraphics[width=\linewidth]{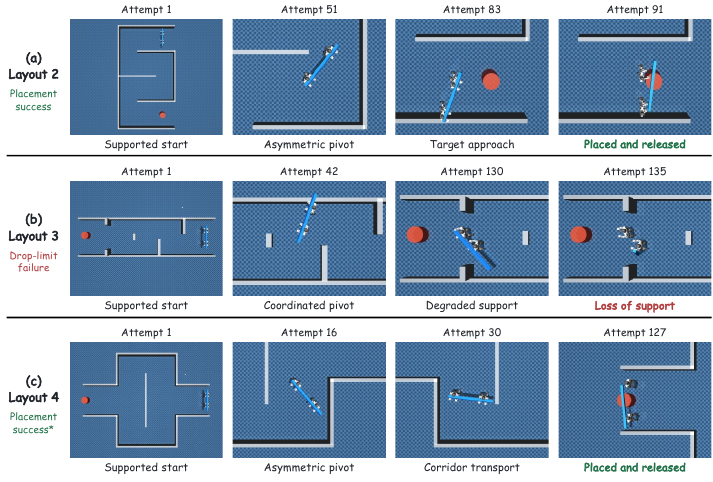}
    \vspace{-2mm}
    \caption{
    \textbf{Representative execution traces across additional layouts.}
    Layout 3 accumulates intermediate progress before payload support
    degrades, illustrating why certified local progress need not preserve
    feasibility for later maneuvers. Layouts 2 and 4 show successful
    placement under different coordination sequences. Attempt labels denote
    internal execution indices rather than cumulative physical transactions.
    }
    \label{fig:multi_cases}
    \vspace{-2mm}
\end{figure}

\paragraph{Local progress does not guarantee future physical feasibility.}
Table~\ref{tab:scene_generalization} records verified placement in Layouts 2
and 4, while Layout 3 reaches no placement despite the highest VP (58.3\%)
and PAUC@160 (53.6\%). The execution traces in Figure~\ref{fig:multi_cases} illustrate this
discrepancy: Layout 3 advances through several certified milestones before
payload support degrades. Because the payload is supported
through finite contact rather than rigid attachment, an accepted transition
can reduce the support margin required by later maneuvers. Checkpoint
acceptance therefore certifies the executed transition, not the feasibility
of the remaining task. The result identifies future support and clearance as
constraints that local transition verification does not explicitly optimize.

The successful Layout 4 continuation uses the refined corridor-verification
condition described in Appendix~\ref{app:scene_execution_records}; it is not
treated as a matched geometry-only replicate.

\subsection{Cross-Model Experience Reuse}
\label{sec:memory_transfer}

Finally, we test whether verified experience can be reused outside the
backbone that produced it. GPT-5.4-mini is initialized with experience
banks collected from successful GPT, Gemini, or Claude executions. No
simulator state, checkpoint, controller history, or trajectory is transferred.

\begin{table*}[t]
    \centering
    \footnotesize
    \setlength{\tabcolsep}{5pt}
    \renewcommand{\arraystretch}{1.10}
    \begin{tabular*}{\textwidth}{
        @{\extracolsep{\fill}}lccccc@{}
    }
        \toprule
        \textbf{Source $\rightarrow$ Target}
        & \shortstack{\textbf{Strategy records /}\\\textbf{action exemplars}}
        & \shortstack{\textbf{Verified}\\\textbf{placement} $\uparrow$}
        & \shortstack{\textbf{VP}\\\textbf{(\%)} $\uparrow$}
        & \shortstack{\textbf{EA}\\\textbf{(\%)} $\uparrow$}
        & \shortstack{\textbf{PAUC@80}\\\textbf{(\%)} $\uparrow$} \\
        \midrule

        GPT $\rightarrow$ GPT
        & 17 / 14
        & 1/5 (20.0\%)
        & 50.0
        & 75.4
        & 38.0 \\

        Gemini $\rightarrow$ GPT
        & 16 / 8
        & 2/5 (40.0\%)
        & 76.7
        & 80.3
        & 55.8 \\

        Claude $\rightarrow$ GPT
        & 4 / 8
        & 0/5 (0.0\%)
        & 56.7
        & $\geq 61.6$
        & 45.0 \\

        \bottomrule
    \end{tabular*}
    \caption{
    Experience reuse with GPT-5.4-mini as the target
    ($n=5$ per source bank). PAUC uses the first 80 cumulative physical
    transactions. $\geq$ denotes the confirmed EA lower bound.
    }
    \label{tab:memory_transfer}
\end{table*}

\paragraph{Bank size alone does not explain cross-model reuse.}
Table~\ref{tab:memory_transfer} records two placements with Gemini-derived
experience, one with GPT-derived experience, and none with the Claude bank.
The target selects, instantiates, and verifies policies, so the experiment
transfers coordination evidence rather than a solved trajectory.
GPT and Gemini provide 17 and 16 strategy records, compared with only four
from Claude, which may reduce the decision coverage available to the target.
Record count is not sufficient, however: GPT has more action exemplars
than Gemini but no pivot exemplar, while both Gemini and Claude include pivot
examples. The observed differences are therefore more consistent with
coverage and state-specific applicability than with bank size alone. Unequal
source banks and the absence of a matched empty-memory condition prevent a
causal transfer claim.

%% file: sections/5_conclusions.tex
\section{Conclusion}
\label{sec:conclusion}

We propose \methodname{}, a framework for adapting executable multi-robot
coordination over fixed motor skills through physical verification,
structured revision, and verified experience. Across three foundation model
families, \methodname{} achieves end-to-end cooperative transport, while
additional evaluations show successful execution under new layouts and
same-model and cross-model experience reuse. The experiments also separate
local execution validity from sustained task advancement: accepted actions
can fail to resolve the current coordination bottleneck, and verified progress
can precede later loss of physical support. These findings identify structured
coordination revision as a useful interface for long-horizon cooperation while
leaving motion precision, contact stability, and physical recovery as open
limitations.

\section{Limitations} 
\label{sec:limitations} 
 
\methodname{} has limitations. First, end-to-end execution incurs 
substantial wall-clock cost because model inference, physical execution, 
verification, and recovery recur throughout a run 
(Appendix~\ref{app:runtime}). Second, rejected executions are recovered by 
restoring an accepted simulator state rather than a physically executable 
recovery policy, which limits direct transfer to real robots. Third, the 
frozen motor interface provides bounded motion intentions rather than precise 
endpoints: unequal robot responses, residual motion after stopping, and 
accumulated formation drift can reduce clearance and destabilize the shared 
payload. Payload stability also depends on finite contact regions, friction, 
and simulator contact dynamics, which we do not systematically randomize in 
the study. Finally, all experiments use simulation, privileged 
physical state information, and supplied route references. Real-world 
deployment will require perception from onboard sensing, more accurate motion 
and contact models, safe physical recovery, and evaluation under broader 
environments and hardware conditions. The relatively small and heterogeneous 
evaluation cohorts also limit strong causal claims about individual 
components, scene generalization, and experience transfer.

\section*{AI Use Statement}

Generative AI tools were used to assist with literature search and
summarization, research ideation and experimental design, software development
and debugging, interpretation of experimental results, and drafting and
refinement of the manuscript. All AI-assisted suggestions, code changes,
experimental configurations, quantitative results, citations, and scientific
claims were independently reviewed and verified by the authors. Generative AI
was not used to generate synthetic datasets or to prove mathematical claims.
The authors take full responsibility for the methodology, implementation,
experiments, analysis, and final content of this work.

\section*{Ethics Statement}

This work studies multi-robot coordination entirely in simulation and does not
involve human participants, personal data, or deployment on physical robots.
The proposed framework is intended for research on cooperative robot
reasoning and execution. Although physical verification and recovery are used
within the simulator, they should not be interpreted as safety guarantees for
real-world deployment. Applying the system to physical robots would require
additional safeguards, state and contact estimation, calibrated motion
control, and hardware-specific safety validation.

\section*{Reproducibility Statement}

Section~\ref{sec:method} specifies the coordination framework.
Appendix~\ref{app:method_details} describes the role interfaces, motor
capabilities, verification, and experience mechanisms;
Appendix~\ref{app:latest_provenance} documents execution settings, baseline
adaptations, endpoints, and selected runs.
Appendix~\ref{app:metric_computation} defines metric reconstruction, uncertainty from unresolved
transactions, resource statistics, and run-time accounting. These descriptions distinguish recorded execution conditions
from the retrospective reporting horizons used in the aggregate results.

%% file: sections/appendix.tex

\clearpage
\appendix

\section{Additional Method Details}
\label{app:method_details}

\subsection{State, Plan, and Role Interfaces}
\label{app:state_and_roles}

The state $s_t=(x_t^{1:N},x_t^o,c_t,\mathcal{G})$ contains robot states,
payload state, contact and support measurements, and scene geometry.
Observations combine serialized measurements, global and payload-centered
views, and geometric references for the active route segment. These
references may include world directions, local progress targets, corridor
boundaries, and completion gates. They are distinct from the semantic plan
$P_t=[(q_k,\gamma_k,\sigma_{k,t})]_{k=1}^{K}$, which records task phases,
completion evidence, and status rather than an executable waypoint sequence.

The tuple $z_t=(\kappa_t,d_t,m_t,\rho_t^{1:N})$ denotes the coordination
fields of the committed subtask. Its structured representation additionally
contains formation intent, completion evidence, and execution constraints.
The Coder supplies bounded numerical parameters $\theta_t^{1:N}$, giving
$\chi_t=(z_t,\theta_t^{1:N})$ and the executable team policy
$\Pi_t=(\pi_t^1,\ldots,\pi_t^N)$. Thus, the Orchestrator determines the
coordination decision, while the Coder determines its numerical and
programmatic implementation.

\begin{table}[htbp]
    \centering
    \small
    \setlength{\tabcolsep}{5pt}
    \renewcommand{\arraystretch}{1.10}
    \begin{tabularx}{\linewidth}{@{}p{0.25\linewidth}X@{}}
        \toprule
        \textbf{Role} & \textbf{Responsibility} \\
        \midrule

        Task Orchestrator
        & Maintain the semantic plan and commit the current team subtask. \\

        Policy Coder
        & Generate bounded parameters and executable policies for the robots. \\

        Evaluator
        & Interpret observations and execution evidence for failure analysis. \\

        Evolution Manager
        & Select a revision scope and objective from verifier evidence. \\

        Mutation Operator
        & Propose a child strategy within the approved scope. \\

        Code Repairer
        & Repair invalid or inconsistent policy drafts. \\

        Memory Writer
        & Propose experience records consistent with executed outcomes. \\

        \bottomrule
    \end{tabularx}
    \caption{
    Structured model roles. Physical acceptance, execution checks,
    checkpoint management, and experience promotion remain host operations.
    }
    \label{tab:app_model_roles}
\end{table}

The roles in Table~\ref{tab:app_model_roles} use the selected backbone
through different prompts and output schemas, not independently trained
models. Their structured proposals remain subject to host validation.

\subsection{Motor Capabilities and Team Execution}
\label{app:skills_and_primitives}
\label{app:coordination_semantics}

The motor interface wraps separate SONIC deployments with frozen policy
weights~\citep{luo2025sonic}. Available skills are summarized in
Table~\ref{tab:frozen_sonic_skills}. The transport primitives are
\texttt{coordinated\_\allowbreak linear\_\allowbreak transport},
\texttt{coordinated\_\allowbreak pivot}, and
\texttt{coordinated\_\allowbreak alignment}.

Linear transport uses a common translation objective; pivoting assigns
temporary inner-support and outer-mover roles; alignment uses a reference
robot while the other makes a bounded correction. Initialization, placement,
release, and retreat are lifecycle operations rather than additional
transport primitives.

\begin{table}[htbp]
    \centering
    \small
    \setlength{\tabcolsep}{4pt}
    \renewcommand{\arraystretch}{1.10}
    \begin{tabularx}{\linewidth}{
        @{}
        >{\raggedright\arraybackslash}p{0.19\linewidth}
        >{\raggedright\arraybackslash}p{0.45\linewidth}
        >{\raggedright\arraybackslash}X
        @{}
    }
        \toprule
        \textbf{Capability}
        & \textbf{Motor skills}
        & \textbf{Arguments} \\
        \midrule

        Longitudinal motion
        & \texttt{carry\_forward}, \texttt{carry\_backward}
        & Distance, speed, ramp duration \\

        Turning
        & \texttt{turn\_clockwise}, \texttt{turn\_counterclockwise}
        & Yaw, yaw rate, turn segment \\

        Lateral motion
        & \texttt{carry\_lateral\_left}, \texttt{carry\_lateral\_right}
        & Distance, speed, ramp duration \\

        Height control
        & \texttt{place\_crouch}, \texttt{recover\_stand}
        & Height, ramp, hold duration \\

        Payload recovery
        & \texttt{recover\_payload\_inward}
        & Ramp and hold duration \\

        \bottomrule
    \end{tabularx}
    \caption{Motor capabilities available through the execution interface.}
    \label{tab:frozen_sonic_skills}
\end{table}

Synchronous coordination uses compatible commands for a shared objective;
asymmetric coordination permits different temporary roles or parameters.
Neither guarantees identical robot displacement. Each robot policy contains
one allowed action call followed by an explicit stop. The host checks output
structure, allowed methods, numerical bounds, syntax, and consistency of the
paired policies before dispatch. Generated policies cannot change the
verifier, invoke simulator restoration, or access low-level controller
handles. Joint dispatch is a logical transaction, not a hard real-time
synchronization barrier.

\subsection{Verification, Revision, and Recovery}
\label{app:verification_and_revision}
\label{app:implementation_boundary}

Acceptance depends on measured payload support and stability, robot state,
environmental contacts, phase validity, and progress relative to the
committed objective. Visual interpretation assists diagnosis but cannot
override contradictory physical measurements. An accepted setup or
corrective action can establish a checkpoint without completing a task
phase; phase advancement requires separate completion evidence.

Revision acts on the failed strategy
$\chi_t=(\kappa_t,d_t,m_t,\rho_t^{1:N},\theta_t^{1:N})$.
Strategy revision concerns the maneuver and objective, coordination revision
concerns mode and temporary roles, and parameter revision concerns bounded
numerical arguments. The host checks the child against its failed parent
and the selected mutable fields. Once approved, that revision constrains the
next orchestration and code-generation step rather than being discarded by
an unrelated decision.

After rejection, the host restores the latest accepted simulator state,
clears transient execution state, resynchronizes the controllers, and
verifies the restoration. A fingerprint of the checkpoint, normalized
policy, and effective arguments blocks identical rejected transactions at
that checkpoint. Recovery neither physically reverses the motion nor
automatically searches earlier checkpoints. Separately selected diagnostic
restart points are evaluation conditions, not autonomous recovery actions.

Algorithm~\ref{alg:coevolve} summarizes this loop. Its model procedures
include bounded validation and repair; an inability to produce a valid next
transaction can terminate execution before the physical budget is exhausted.
Only dispatched physical transactions consume that budget. Execution
attempt identifiers can therefore differ from cumulative transaction counts.

\subsection{Experience Contents and Retrieval}
\label{app:memory_retrieval}

The two dynamic stores are written as
$\mathcal{M}_t=(\mathcal{M}_t^D,\mathcal{M}_t^A)$.
The static skill catalog belongs to the capability interface and does not
evolve. Strategy records store applicability conditions, coordination
principles, observed failure patterns, and repair directions. The active
phase filters applicable records, while recent failure type and level
contribute to ranking. The next primitive, objective, and mode are not query
conditions because they have not yet been selected.

After the coordination decision is committed, action exemplars are matched
by primitive, semantic objective, and coordination mode. Phase only ranks
otherwise compatible exemplars. These entries retain accepted code and
effective parameters but remain subject to the current subtask and execution
constraints. Strategy retrieval therefore informs the coordination decision;
exemplar retrieval informs its implementation.

Promotion requires agreement between a record's claim, the executed behavior,
and the verifier outcome. Eligible accepted executions can yield positive
strategy records and executable exemplars; rejected executions can yield
counterexamples. Unexecuted repairs are not successful experience. Strategy
records omit scene coordinates, checkpoint poses, and numerical action
magnitudes, whereas exemplars retain concrete implementations. Standard
fresh runs start with empty dynamic memory; transfer runs preload a fixed
source bank and update only their own subsequent experience.

\subsection{Representative Coordination Revision}
\label{app:qualitative_revision}

\begin{figure}[htbp]
    \centering
    \includegraphics[width=\linewidth]{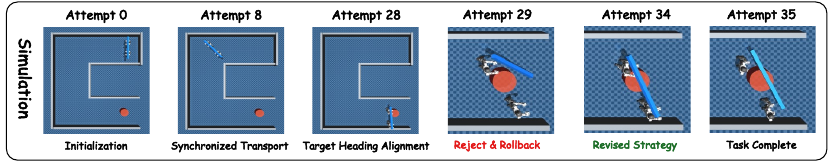}
    \caption{
    \textbf{Coordination revision during execution.}
    Selected snapshots show transport, target alignment, rejection and
    checkpoint restoration, followed by revised behavior and the illustrated
    completion. Attempt labels identify positions in the episode, not
    cumulative physical-transaction counts.
    }
    \label{fig:case_study}
\end{figure}

Figure~\ref{fig:case_study} illustrates a rejection, restoration, and
subsequent coordination revision within an episode. It complements the
cross-layout visualization in Figure~\ref{fig:multi_cases}; neither figure
adds independent runs to the quantitative evaluation.

\section{Execution and Evaluation Details}
\label{app:latest_provenance}

\subsection{Environment Layouts and Endpoints}
\label{app:generalization_scenes}
\label{app:evaluation_records}

\begin{figure}[htbp]
    \centering
    \includegraphics[width=\linewidth]{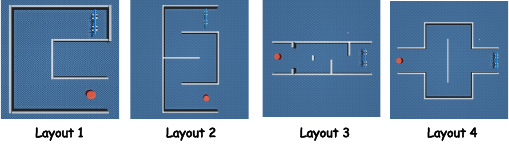}
    \caption{
    \textbf{Environment layouts.}
    Layout 1 is the base environment; Layouts 2--4 provide additional
    corridor configurations.
    }
    \label{fig:generalization_scenes}
    \label{fig:layout_overview}
\end{figure}

The base environment and three additional corridor layouts are shown in
Figure~\ref{fig:generalization_scenes}. All use two humanoids supporting a
free rigid payload. The images describe scene geometry, not a prescribed
executable trajectory. Route references, native milestones, and verification
predicates are associated with the evaluated scene configuration.

The base task requires stable placement, release, and robot retreat. The
scene and experience evaluations omit retreat from their placement endpoint.
Placement requires target support and release over a $3\,\mathrm{s}$
stability window. The reported base tolerances include a target support
radius of $0.30\,\mathrm{m}$, payload-center vertical tolerance of
$0.08\,\mathrm{m}$, and terminal tilt tolerance of $20^\circ$.
The full-task endpoint additionally requires upright robots and the
specified robot--target clearance.

\subsection{Comparison Configurations}
\label{app:baseline_configurations}

All comparison conditions use GPT-5.4-mini. REMAC~\citep{yuan2025remac},
COHERENT~\citep{liu2025coherent}, CaP-Agent0~\citep{fu2026capx}, and
ASPIRE~\citep{lu2026aspire} are adapted to the cooperative transport task
rather than reproduced in their original environments.
Table~\ref{tab:baseline_adaptation} describes the evaluated mechanisms.
A shared task interface does not imply identical runtime or recovery
implementations.

\begin{table}[htbp]
    \centering
    \small
    \setlength{\tabcolsep}{5pt}
    \renewcommand{\arraystretch}{1.10}
    \begin{tabularx}{\linewidth}{@{}p{0.31\linewidth}X@{}}
        \toprule
        \textbf{Condition} & \textbf{Evaluated mechanism} \\
        \midrule

        REMAC
        & Reflective multi-robot planning through the task and capability
          interface. \\

        COHERENT
        & Multi-agent coordination using the robot observations and bounded
          skills. \\

        CaP-Agent0
        & Executable robot-code generation for cooperative payload transport. \\

        ASPIRE (Online)
        & Online adaptation with the reported setting $K=8$ and verification
          local to the current checkpoint. \\

        Flat Reflective Repair
        & Direct policy rewriting without selecting a typed revision scope;
          physical feedback, recovery, and dynamic memory remain available. \\

        w/o Dynamic Experience
        & Structured revision with strategy and exemplar retrieval, writing,
          and promotion disabled. \\

        \bottomrule
    \end{tabularx}
    \caption{
    Comparison mechanisms. External methods are task-specific adaptations;
    the component conditions alter the stated parts of \methodname{}.
    }
    \label{tab:baseline_adaptation}
\end{table}

The full reference in Table~\ref{tab:baselines_ablations} reuses the GPT
runs in Table~\ref{tab:backbone_main}. Differences in execution
configuration and repair continuations prevent interpreting this comparison
as an isolated causal estimate of a component's effect.

\subsection{Controller and Contact Configuration}
\label{app:sonic_execution_details}
\label{app:contact_model}

The execution chain consists of skill wrappers, batched planner intents,
two independent SONIC processes, separate robot command streams, and the
dual-robot MuJoCo bridge. The inspected deployment uses nominal input,
planner, policy/control, and command-writer rates of 100, 10, 50, and
500 Hz, respectively. These are scheduling settings, not guarantees of
simultaneous command consumption. The inspected manifest records MuJoCo
3.8.1; the separate stopping diagnostic uses a $0.005\,\mathrm{s}$
integration step.

Each deployed policy uses ten-frame local histories containing joint states,
base angular velocity, gravity direction, and previous actions, together
with motion references. The robots maintain independent histories and
reference states. The inspected support adapter regulates 14 arm joints per
robot using torque-limited PD control with $k_p=24$, $k_d=1.5$, a
$6\,\mathrm{s}$ ramp, and palms tilted by $60^\circ$; the waist is
excluded. This posture regulation is not a rigid grasp or a constraint on
the payload center of mass.

The palm proxy has half extents of $0.06\,\mathrm{m}$,
$0.035\,\mathrm{m}$, and $0.006\,\mathrm{m}$; the corresponding forearm
values are $0.11\,\mathrm{m}$, $0.032\,\mathrm{m}$, and
$0.006\,\mathrm{m}$. Contact parameters from the inspected generator are
listed in Table~\ref{tab:contact_parameters}.
They are simulation parameters rather than measured material coefficients;
effective contact also depends on geometry, mixing rules, and the solver.
They are not a claim that every historical instantiated scene was identical.

\begin{table}[htbp]
    \centering
    \small
    \setlength{\tabcolsep}{8pt}
    \begin{tabular}{@{}lccc@{}}
        \toprule
        \textbf{Geometry}
        & \textbf{Sliding}
        & \textbf{Torsional}
        & \textbf{Rolling} \\
        \midrule

        Palm/forearm proxies & 3.0 & 0.05 & 0.002 \\
        Payload              & 2.0 & 0.05 & 0.002 \\
        Target               & 1.4 & 0.03 & 0.001 \\
        Ground default       & 1.0 & 0.02 & 0.001 \\

        \bottomrule
    \end{tabular}
    \caption{Configured contact parameters in the inspected scene generator.}
    \label{tab:contact_parameters}
\end{table}

For payload orientation $R$, center height $z_o$, and half extents $h_i$,
its lowest point is $z_{\min}=z_o-\sum_{i=1}^{3}|R_{3i}|h_i$.
This calculation addresses payload clearance only; it does not certify robot
clearance or the full swept trajectory. No contact-parameter sensitivity or
randomization study is included in the reported evaluation.

\subsection{Local Execution Diagnostics}
\label{app:stopping_diagnostic}
\label{app:release_feedback}

\paragraph{Stopping response.}
A local test without foundation-model calls requests a $0.25\,\mathrm{m}$
translation using the native $0.5\,\mathrm{m/s}$ speed and
$1.5\,\mathrm{s}$ ramp. Displacements are projected onto the fixed initial
world direction, with an endpoint tolerance of $\pm0.05\,\mathrm{m}$.
Table~\ref{tab:native_stopping_diagnostic} reports this single trace.

\begin{table}[htbp]
    \centering
    \small
    \setlength{\tabcolsep}{6pt}
    \renewcommand{\arraystretch}{1.08}
    \begin{tabular}{@{}lr@{}}
        \toprule
        \textbf{Quantity} & \textbf{Observed value} \\
        \midrule

        Requested displacement
        & $0.250\,\mathrm{m}$ \\

        Payload displacement at stop trigger
        & $0.192672\,\mathrm{m}$ \\

        Final payload displacement
        & $0.663442\,\mathrm{m}$ \\

        Maximum sampled payload displacement
        & $0.689713\,\mathrm{m}$ \\

        Payload displacement after stop trigger
        & $0.470769\,\mathrm{m}$ \\

        Front robot displacement after stop trigger
        & $0.476529\,\mathrm{m}$ \\

        Back robot displacement after stop trigger
        & $0.483128\,\mathrm{m}$ \\

        Endpoint error
        & $+0.413442\,\mathrm{m}$ \\

        \bottomrule
    \end{tabular}
    \caption{
    Single-trace stopping diagnostic, not a repeated-trial accuracy estimate.
    }
    \label{tab:native_stopping_diagnostic}
\end{table}

Bilateral IDLE sending completes approximately $0.742\,\mathrm{ms}$ after
the trigger, with an observed consumption bound of approximately
$13.7\,\mathrm{ms}$ after sending. Commands remain IDLE while references
and both robots continue forward. The trace documents residual motion after
stop consumption without uniquely identifying its lower-level cause.
Successful return from a skill call therefore does not establish endpoint
accuracy or continuous-time safety.

\paragraph{Release feedback.}
An earlier execution fails a pre-release vertical condition at a logged
$0.459\,\mathrm{m}$ against a $0.45\,\mathrm{m}$ limit. Support withdrawal
does not occur. The feedback is subsequently made more specific about the
stage, predicate, measurement, and withdrawal status without relaxing the
gate. The missing historical gate snapshot is not replaced with a later
measurement. This is a feedback correction, not evidence of a post-release
drop or repaired release dynamics.

\subsection{Scene Records and Execution Conditions}
\label{app:method_version_boundary}
\label{app:scene_execution_records}

Execution changes address approved child strategies, release feedback, route
boundaries, accepted setup history, and restored observations. These do not
replace the conceptual method, but can affect the subsequent execution.
The successful Layout 4 continuation additionally uses adjusted corridor
verification. Normal rollback still restores the latest accepted checkpoint;
separately authorized earlier restart points are not automatic ancestor
search. Original budgets and drop rules remain part of the corresponding
records.

\begin{table}[htbp]
    \centering
    \small
    \setlength{\tabcolsep}{5pt}
    \renewcommand{\arraystretch}{1.10}
    \begin{tabular*}{\linewidth}{
        @{\extracolsep{\fill}}lccc@{}
    }
        \toprule
        \textbf{Layout}
        & \textbf{Run 1}
        & \textbf{Run 2}
        & \textbf{Run 3} \\
        \midrule

        Layout 2
        & D, $19/80$
        & S, $88/160$
        & D, $30/80$ \\

        Layout 3
        & B, $160/160$
        & D, $118/160$
        & D, $94/160^{\dagger}$ \\

        Layout 4
        & D, $57/80$
        & D, $29/80$
        & S, $120/160^{\ast}$ \\

        \bottomrule
    \end{tabular*}
    \caption{
    Scene outcomes and cumulative transactions/original limit.
    S denotes placement, D a drop-limit terminal, and B budget exhaustion.
    $^{\ast}$ Placement follows adjusted corridor verification.
    $^{\dagger}$ The effective limit is 160; an older embedded field retains 80.
    }
    \label{tab:scene_execution_records}
\end{table}

Layout 2 and Layout 4 reach placement at 88 and 120 transactions.
Neither is a placement within 80 transactions. The common PAUC horizon of
160 is a retrospective accounting window; it does not recast originally
shorter executions as fresh trials with identical budgets.

\subsection{Source Banks and Transfer Records}
\label{app:transfer_execution_records}

The GPT, Gemini, and Claude banks contain 17/14, 16/8, and 4/8
strategy-record/action-exemplar entries, respectively. Each fixed bank
initializes five target executions. A successful source can contribute both
positive strategy records and counterexamples; its action exemplars are
accepted implementations. No source physical state or complete trajectory
is transferred. Unequal bank sizes and absent matched empty initialization
limit conclusions about transfer benefits.

\begin{table}[htbp]
    \centering
    \footnotesize
    \setlength{\tabcolsep}{3pt}
    \renewcommand{\arraystretch}{1.12}
    \begin{tabular*}{\linewidth}{
        @{\extracolsep{\fill}}lccccc@{}
    }
        \toprule
        \textbf{Source}
        & \textbf{Run 1}
        & \textbf{Run 2}
        & \textbf{Run 3}
        & \textbf{Run 4}
        & \textbf{Run 5} \\
        \midrule

        GPT
        & D, $25/80$
        & S$^{\dagger}$, $56/80$
        & D, $46/80$
        & G, $25/80$
        & D, $37/80$ \\

        Gemini
        & D, $40/80$
        & S$^{\dagger}$, $58/80$
        & C, $82/100$
        & S, $41/80$
        & G, $46/80$ \\

        Claude
        & D, $56/80$
        & B$^{\dagger}$, $80/80$
        & D$^{\dagger}$, $53/80$
        & D$^{\ddagger}$, $32/80$
        & D, $78/80$ \\

        \bottomrule
    \end{tabular*}
    \caption{
    Transfer outcomes and cumulative transactions/original limit.
    G denotes generation/repair stall; C denotes conditional diagnostic stop.
    $^{\dagger}$ Original finalization gaps are retained alongside later
    read-only validation.
    $^{\ddagger}$ The terminal outcome is independently reconciled.
    }
    \label{tab:transfer_execution_records}
\end{table}

The records in Table~\ref{tab:transfer_execution_records} include repair
continuations and an authorized restart in the Claude condition. Read-only
validation confirms the retained records without replacing the original
finalization status or creating new physical trials. The Gemini conditional
stop at 82/100 is not reclassified as generation failure or budget
exhaustion. For the fourth Claude execution, original metrics record 31
transactions before a camera exception; independent reconciliation confirms
32 transactions and the sixth drop. The original exception remains recorded.
Two charged outcomes in the second Claude execution remain unresolved.

\paragraph{Checkpoint consistency of decision context.}
A Claude-to-GPT transfer execution illustrates a distinction between restoring
physical state and restoring a valid decision context. After simulator
recovery, the Manager continued using a decision associated with a retired
checkpoint. Correcting the context binding allowed execution to resume, but
the run ultimately reached the payload-drop limit. This separates two
requirements: plans and retrieved evidence must apply to the state from which
a policy is executed, and the resulting motion must preserve physical support.
Repairing the former does not establish the latter. The case motivates
checking checkpoint consistency when reusing decisions after recovery rather
than treating simulator restoration alone as sufficient synchronization.

\section{Metric Computation and Supplementary Results}
\label{app:metric_computation}

\subsection{Outcomes, Progress, and Acceptance}
\label{app:metric_definitions}

All group metrics weight runs equally. Success is
$\mathrm{SR}=100n^{-1}\sum_r S_r$, using full-task completion within 80
transactions for the base tables and placement/release for the scene and
transfer tables. Verified progress is
$\mathrm{VP}=100n^{-1}\sum_r q_r/K_r$, with $q_r$ certified milestones
in the terminal trusted-state record and $K_r$ native scene milestones.
The base task has four route gates, a target-region milestone, and a
protocol-specific terminal milestone. Milestone fractions are not fractions
of distance, time, or skill difficulty.

For executions with $T_r>0$ charged physical transactions and $A_r$
confirmed acceptances, $\mathrm{EA}=100n^{-1}\sum_r A_r/T_r$.
This is the mean of per-run ratios, not $100\sum_r A_r/\sum_r T_r$.
If $U_r$ charged outcomes are unresolved, the accounting bounds are
\[
\mathrm{EA}_{\mathrm{lo}}=
\frac{100}{n}\sum_r\frac{A_r}{T_r},
\qquad
\mathrm{EA}_{\mathrm{hi}}=
\frac{100}{n}\sum_r\frac{A_r+U_r}{T_r}.
\]
The main tables report only the confirmed lower endpoint with $\geq$.
Unresolved outcomes are not relabeled as rejections. This yields the
reported lower bounds of 83.6\% for COHERENT and 61.6\% for the Claude
source condition. These are bookkeeping bounds rather than confidence
intervals. Setup and corrective actions may be accepted without completing
a milestone, so EA describes local execution validity, not global success.

\subsection{Reconstructing Progress AUC}
\label{app:pauc_reconstruction}

PAUC is computed retrospectively from saved accepted checkpoints. For each
run, records are ordered by cumulative charged physical transactions,
including inherited continuation costs. The first saved checkpoint that
certifies a prefix containing milestone $k$ determines $\tau_{rk}$;
$\tau_{rk}=\infty$ when no certificate is recorded. We use
$p_r(t)=K_r^{-1}\sum_k\mathbf{1}[\tau_{rk}\leq t]$ with $p_r(0)=0$.
For an integer reporting horizon $B$,
\[
\begin{aligned}
\mathrm{PAUC@}B
&=\frac{100}{nB}\sum_r\int_0^B p_r(t)\,dt \\
&=\frac{100}{nB}\sum_r\sum_{t=0}^{B-1}p_r(t) \\
&=\frac{100}{n}\sum_r\frac{1}{K_rB}
  \sum_{k=1}^{K_r}\left[B-\min(\tau_{rk},B)\right].
\end{aligned}
\]
The integral and integer sum use the same step-function convention.
Attainment exactly at $B$ contributes zero area, although it can still
establish endpoint success. A run with no certified milestones scores zero.

Each milestone earns credit once. Rejected transactions and unresolved
charged outcomes remain in the chronology but receive no unverified
progress credit. Setup-only acceptance and repeated attainment introduce
no additional milestone credit. Rollback does not refund transactions;
a certified prefix later retired by a documented rewind retains its
historical first-attainment credit. Unverified fragments from different
branches are not combined into a certificate. After termination, recorded
attainment is held constant to $B$.

PAUC consequently measures historical attainment and its timing, while VP
measures terminal retained progress. Neither substitutes for success or
continuous physical stability. For example, attaining half the milestones
at transaction 20 under $B=80$, with no further progress, gives
$100(1/2)(80-20)/80=37.5\%$; attaining the same milestones at transaction
60 gives 12.5\%. These are illustrative scores, not additional experiment
results or percentages of resource savings.

\begin{table}[htbp]
    \centering
    \small
    \setlength{\tabcolsep}{5pt}
    \renewcommand{\arraystretch}{1.10}
    \begin{tabularx}{\linewidth}{@{}p{0.26\linewidth}cX@{}}
        \toprule
        \textbf{Evaluation}
        & \textbf{PAUC horizon}
        & \textbf{Outcome scope} \\
        \midrule

        Base task
        & 80
        & Placement, release, and retreat within 80 transactions. \\

        Additional layouts
        & 160
        & Placement/release in the selected execution; original limits differ. \\

        Experience reuse
        & 80
        & PAUC uses the first 80 transactions; other metrics use the complete
          selected execution. \\

        \bottomrule
    \end{tabularx}
    \caption{
    Reporting horizons and task endpoints. A common retrospective horizon
    does not establish a matched experimental protocol.
    }
    \label{tab:metric_horizons}
\end{table}

The score retains the native verification semantics of the saved records.
It is not obtained by replaying all runs under one common verifier.
Different milestone definitions or execution configurations limit direct
causal comparisons even when all scores are expressed as percentages.

\subsection{Resource Use and Termination Categories}
\label{app:execution_diagnostics}

Table~\ref{tab:execution_diagnostics} reports actual physical interaction
and recorded foundation-model calls, including continuation costs once.
Pre-dispatch failures consume model calls but no physical transactions.
Early unsuccessful termination can reduce both counts, so neither has an
unconditional better-direction arrow. Incomplete call totals remain lower
bounds and are not converted into monetary cost without billing evidence.

Termination labels describe the final stopping condition, not intermediate
error counts or unique causes. D is the applicable payload-drop limit;
later placement records use six cumulative drops, while older records retain
their configured stopping rule. G is generation/repair stall, covering the
failure to produce an executable next candidate through invalid,
contradictory, or repeatedly rejected proposals. It does not distinguish
model errors from interface or recovery-rule effects. B is exhaustion of the
physical budget and C is a conditional diagnostic stop.

\begin{table}[htbp]
    \centering
    \footnotesize
    \setlength{\tabcolsep}{4pt}
    \renewcommand{\arraystretch}{1.10}
    \begin{tabularx}{\linewidth}{
        @{}>{\raggedright\arraybackslash}Xcccc@{}
    }
        \toprule
        \textbf{Condition}
        & \shortstack{\textbf{Physical}\\\textbf{transactions}}
        & \shortstack{\textbf{FM}\\\textbf{calls}}
        & \shortstack{\textbf{Terminal counts}\\\textbf{D/G/B/C}}
        & \shortstack{\textbf{Unresolved}\\\textbf{transactions}} \\
        \midrule

        \multicolumn{5}{@{}l}{
        \emph{Base-task backbones (five runs each)}} \\

        GPT-5.4-mini
        & 52.4 & 460.4 & 0/1/1/0 & 0 \\

        Gemini-3.5-Flash
        & 27.0 & 148.6 & 0/2/0/0 & 0 \\

        Claude-Sonnet-4.6
        & 24.2 & 182.0 & 0/3/0/0 & 0 \\

        \addlinespace[3pt]
        \multicolumn{5}{@{}l}{
        \emph{Comparison conditions (five runs each)}} \\

        REMAC
        & 20.4 & 122.8 & 4/1/0/0 & 0 \\

        COHERENT
        & 43.8 & 873.4 & 1/4/0/0 & 1 \\

        CaP-Agent0
        & 17.4 & $\geq355.0$ & 1/4/0/0 & 0 \\

        ASPIRE (Online)
        & 19.2 & $\geq175.2$ & 1/4/0/0 & 0 \\

        Flat Reflective Repair
        & 29.6 & 176.6 & 4/1/0/0 & 0 \\

        w/o Dynamic Experience
        & 62.2 & 490.6 & 0/4/1/0 & 0 \\

        \addlinespace[3pt]
        \multicolumn{5}{@{}l}{
        \emph{Additional layouts (three runs each)}} \\

        Layout 2
        & 45.7 & 301.7 & 2/0/0/0 & 0 \\

        Layout 3
        & 124.0 & $\geq1394.7$ & 2/0/1/0 & 0 \\

        Layout 4
        & 68.7 & 650.7 & 2/0/0/0 & 0 \\

        \addlinespace[3pt]
        \multicolumn{5}{@{}l}{
        \emph{Source experience $\rightarrow$ GPT (five runs each)}} \\

        GPT $\rightarrow$ GPT
        & 37.8 & 282.8 & 3/1/0/0 & 0 \\

        Gemini $\rightarrow$ GPT
        & 53.4 & 348.6 & 1/1/0/1 & 0 \\

        Claude $\rightarrow$ GPT
        & 59.8 & $\geq576.2$ & $4^{\ddagger}$/0/1/0 & 2 \\

        \bottomrule
    \end{tabularx}
    \caption{
    Resource and termination diagnostics. Physical transactions and FM calls
    are per-run means; terminal and unresolved counts are group totals.
    $\geq$ denotes a confirmed call-count lower bound.
    $^{\ddagger}$ One drop terminal is independently reconciled with its
    original infrastructure exception retained. The full comparison row
    reuses the GPT backbone results and is not duplicated here.
    }
    \label{tab:execution_diagnostics}
\end{table}

The selected archive contains 69 logical executions after deduplication:
15 backbone runs, 30 comparison runs, nine scene runs, and 15 transfer runs.
Reused reference rows and continuation prefixes are not additional
independent trials. Different endpoints, limits, and execution conditions
preclude combining this archive into one aggregate success rate.

\subsection{Run-Time Accounting}
\label{app:runtime}

Run time is measured as elapsed wall-clock time within recorded execution
segments, summed once across all segments belonging to each logical execution.
It includes model and API waiting, physical execution, verification, recovery,
and finalization. Manual pauses, code modification, and queue waiting between
segments are excluded, and concurrent subprocess durations within a segment
are not summed separately.

The selected archive contains 69 deduplicated logical executions. Timing is
complete for 65 executions; the remaining four retain confirmed lower bounds
from available continuous event spans. The resulting cumulative run time is
at least $112.8$ hours, with a mean of at least $98.1$ minutes per execution.
The Full condition in Table~\ref{tab:baselines_ablations} reuses the five GPT
executions in Table~\ref{tab:backbone_main} and is not counted again. The
incomplete timing records are COHERENT run 203, Layout~3 runs 200 and 201,
and Claude-to-GPT run 203. Missing intervals are not imputed as zero.

\begin{table*}[t]
    \centering
    \footnotesize
    \setlength{\tabcolsep}{6pt}
    \renewcommand{\arraystretch}{1.08}
    \begin{tabular}{lccr}
        \toprule
        \textbf{Condition}
        & \textbf{$n$}
        & \textbf{Complete timing}
        & \textbf{Mean run time (min)} \\
        \midrule

        \multicolumn{4}{l}{\emph{Base-task backbones}} \\
        GPT-5.4-mini
        & 5 & 5/5 & 254.7 \\
        Gemini-3.5-Flash
        & 5 & 5/5 & 74.9 \\
        Claude-Sonnet-4.6
        & 5 & 5/5 & 87.8 \\

        \addlinespace
        \multicolumn{4}{l}{\emph{Comparison conditions}} \\
        REMAC
        & 5 & 5/5 & 29.0 \\
        COHERENT
        & 5 & 4/5 & $\geq 99.3$ \\
        CaP-Agent0
        & 5 & 5/5 & 29.6 \\
        ASPIRE (Online)
        & 5 & 5/5 & 55.6 \\
        Flat Reflective Repair
        & 5 & 5/5 & 38.7 \\
        w/o Dynamic Experience
        & 5 & 5/5 & 158.6 \\

        \addlinespace
        \multicolumn{4}{l}{\emph{Additional layouts}} \\
        Layout 2
        & 3 & 3/3 & 74.5 \\
        Layout 3
        & 3 & 1/3 & $\geq 253.0$ \\
        Layout 4
        & 3 & 3/3 & 138.3 \\

        \addlinespace
        \multicolumn{4}{l}{\emph{Source experience $\rightarrow$ GPT}} \\
        GPT $\rightarrow$ GPT
        & 5 & 5/5 & 57.2 \\
        Gemini $\rightarrow$ GPT
        & 5 & 5/5 & 81.2 \\
        Claude $\rightarrow$ GPT
        & 5 & 4/5 & $\geq 108.3$ \\

        \midrule
        \textbf{All selected executions}
        & \textbf{69}
        & \textbf{65/69}
        & $\boldsymbol{\geq 98.1}$ \\

        \bottomrule
    \end{tabular}
    \caption{
    Run-time statistics for the selected execution archive. Means include
    successful and unsuccessful outcomes. Complete timing counts executions
    with complete recorded timing; $\geq$ denotes a confirmed lower bound
    caused by missing historical timing intervals. The reused Full condition
    from Table~\ref{tab:baselines_ablations} is not counted separately.
    }
    \label{tab:runtime}
\end{table*}

Run time describes end-to-end execution-process duration rather than robot
motion time, GPU-hours, or the calendar duration of the evaluation campaign.
It covers only the selected evaluation archive, not development experiments,
manual code changes, or diagnostic work outside these records. Shorter run
time does not necessarily indicate greater efficiency because unsuccessful
runs may terminate early, while repeated verification and recovery can extend
a run substantially.

Mean run time over all executions should also be distinguished from run time
conditioned on success. The five GPT base-task executions average
$254.7$ minutes, whereas the three successful GPT executions average
$142.4$ minutes. One budget-exhausted GPT run lasts approximately
$11.65$ hours and substantially increases the all-run mean. We therefore
report all-run mean time as an end-to-end execution statistic rather than as
a better-is-lower efficiency metric.